\documentclass[a4paper,twoside]{article}

\usepackage{epstopdf}
\usepackage{multirow}
\usepackage{url}
\usepackage[all]{nowidow}
\usepackage[nolist]{acronym}
\usepackage{tikz}

\usepackage{epsfig}
\usepackage{subcaption}
\usepackage{calc}
\usepackage{amssymb}
\usepackage{amstext}
\usepackage{amsmath}
\usepackage{amsthm}
\usepackage{multicol}
\usepackage{pslatex}
\usepackage{apalike}
\usepackage{SCITEPRESS}     

\begin{acronym}[BRSM]
 \acro{GI}{Global Illumination}
 \acro{CD}{Chamfer Distance}
 \acro{EMD}{Earth  Mover’s  Distance}
 \acro{HD}{Hausdorff Distance}
 \acro{LOP}{Locally Optimal Projection}
 \acro{GAN}{Generative Adversarial Network}
 \acro{MLP}{Multi Layer Perceptron}
 \acro{TSDF}{Truncated Signed Distance Function}
\end{acronym}

\usetikzlibrary{plotmarks}
\usetikzlibrary{arrows.meta}
\usetikzlibrary{arrows}
\usetikzlibrary{spy}
\usetikzlibrary{calc}
\usetikzlibrary{shapes}
\usetikzlibrary{fit}
\usetikzlibrary{positioning}
\usetikzlibrary{3d}
\usetikzlibrary{quotes}

\begin{document}

\title{Point Cloud Upsampling and Normal Estimation using Deep Learning for Robust Surface Reconstruction} 



\author{\authorname{Rajat Sharma\sup{1}, Tobias Schwandt\sup{1}, Christian Kunert\sup{1}, Steffen Urban\sup{2} and Wolfgang Broll\sup{1}}
\affiliation{
\sup{1}Ilmenau University of Technology, Virtual Worlds and Digital Games Group, Ehrenbergstra\ss e 29, Ilmenau, Germany
}
\affiliation{
\sup{2}Carl Zeiss AG, Corporate Research and Technology, Carl-Zeiss-Promenade 10, Jena, Germany
}
\email{
\{rajat.sharma, tobias.schwandt, christian.kunert\}@tu-ilmenau.de, steffen.urban@zeiss.com, wolfgang.broll@tu-ilmenau.de
}
}


\abstract{The reconstruction of real-world surfaces is on high demand in various applications.
Most existing reconstruction approaches apply 3D scanners for creating point clouds which are generally sparse and of low density.
These points clouds will be triangulated and used for visualization in combination with surface normals estimated by geometrical approaches.
However, the quality of the reconstruction depends on the density of the point cloud and the estimation of the surface normals.
In this paper, we present a novel deep learning architecture for point cloud upsampling that enables subsequent stable and smooth surface reconstruction. 
A noisy point cloud of low density with corresponding point normals is used to estimate a point cloud with higher density and appendant point normals. 
To this end, we propose a compound loss function that encourages the network to estimate points that lie on a surface including normals accurately predicting the orientation of the surface. 
Our results show the benefit of estimating normals together with point positions. 
The resulting point cloud is smoother, more complete, and the final surface reconstruction is much closer to ground truth.}

\onecolumn \maketitle \normalsize \setcounter{footnote}{0} \vfill

\section{\uppercase{Introduction}}
\label{sec:introduction}

\noindent Point clouds are one special representation of objects in 3D space. 
The recent availability of large range 3D scanner devices has led to the research and development in many areas of point cloud processing \cite{weinmann2016reconstruction}. 
Classically, large amounts of handcrafted features were extracted from the point clouds (e.g., eigenvalues or entropy), making sophisticated neighborhood and feature selection necessary. Although in the area of deep learning, there has not been much done in the early years.
But since the adaption of PointNet \cite{qi2016pointnet} and PointNet++ \cite{qi2017pointnetplusplus} much work has been done on point clouds. 
These deep networks are trained to alleviate the process of feature engineering and selection and have done very well in the area of object classification and segmentation. 
The feature extraction and classification accuracy is further improved with new architectures like Graph-CNNs \cite{ZhangR_18_gcnn_point_cloud} using graph convolution and Dynamic Graph-CNNs \cite{dgcnn} which also extracts edge features using edge convolution. 

In the case of deep learning, the representation of a 3D object in space is very important as this basically determines the network structure as well as input and output representations. 
In case of volumetric representation, the network is applied to a 3D voxel grid structure demonstrated in VoxNet \cite{maturana2015voxnet}, OctnetFusion \cite{Riegler2017OctNet}.

In our work, we use point based representation to perform point cloud densification. 
Approaches such as PU-Net \cite{Yu18}, EC-Net \cite{yu2018ec} etc. considers the problem of upsampling the point clouds.
All of these approaches consider only point cloud as output but we also estimate the normals by upsampling. 
Surface normals are also an important property of point clouds. 
It has applications in surface reconstruction, rendering as well as lighting calculations of 3D model which motivates us to estimate point cloud with normals.
We compare our results with some of the mentioned approaches in this paper.
In conclusion our main contributions are:

\begin{itemize}
  \item the extraction of features from 3D point cloud patches to upsample a input point cloud using feature reshaping. 
  \item being one of the first approaches for an accurate prediction of surface normals along with upsampled point clouds using a neural network.
  \item a novel loss function consisting of point and normal loss for a more accurate upsampling.
\end{itemize}

This paper is structured as follows: in section \ref{sec:introduction}, a brief introduction into the research topic is provided with some specific related work in section \ref{sec:related_work}.
Subsequently in section \ref{sec:network_architecture}, the network architecture as well as the feature embedding and reshaping is emphasized.
Section \ref{sec:loss_function} describes the changes of the loss function made to the architecture.
The result of the new provided approach is shown in section \ref{sec:evaluation} in comparison to other related approaches.
Inside section \ref{sec:discussion}, the findings and limitations are discussed.
Next, all findings are concluded in section \ref{sec:conclusion} with a brief discussion and some possible future work.


\section{\uppercase{Related Work}} \label{sec:related_work}

\noindent In this section, we will review existing approaches based on upsampling of point clouds and how they are different from our approach. 
After discussing classic optimization based methods, we will then look into methods based on deep learning.

Early work on denoising point clouds is done in Parameterization-free Projection for Geometry Reconstruction \cite{Lipman2007Parameterization-freeReconstruction} which uses the \ac{LOP} operator for surface prediction. 
In addition, this method introduces a regularization term to move the point near to the surface. 
\cite{Huang2009ConsolidationReconstruction} proposes weighted \acs{LOP} (WLOP) to get a denoised and evenly distributed point cloud. 
In addition, it also estimates the point normals.
Since these methods work well but fail at very sharp edges, \cite{Huang2013Edge-awareResampling} introduces edge aware resampling to preserve the features at sharp edges. 
It first resamples points away from edges and calculate the normals. 
Based on these reliable points and normals, it progressively resamples towards the edges. 
L0 Minimization \cite{Sun2015DenoisingMinimization} takes point cloud and normals as input and resamples it such that all point lies on the surface with correct normals. 

Initially, volumetric approaches were used to make high resolution voxel grids using deep learning for accurate surface reconstruction. VoxNet \cite{maturana2015voxnet} uses 3D convolution on voxel grids for the classification tasks but the approach is limited to a \( 32^3 \) resolution. 
However, the Octnet \cite{Riegler2017OctNet} approach utilizes the convolution operation up to a level of \( 256^3 \) using space partitioning by converting 3D voxel grids into small octree structures. 
Based on Octnet, OctnetFusion \cite{Riegler2017OctNetFusion} takes low resolution \ac{TSDF} volume and estimates higher resolution \ac{TSDF} volume. 
Similarly, other approaches like \cite{hspHane19} also use octrees instead of 3D volumes.
Depending on this concrete data structure, these approaches suffer from performance issue like memory and timing requirements when increasing the input resolution.

In the point cloud domain, PU-Net \cite{Yu18} introduced a first point upsampling network. 
They divide the mesh into small patches and group the point w.r.t. different radii values and extract features. 
Subsequently, all features are concatenated and reshaped to yield upscaled point features. 
Based on this architecture, they introduced another network that preserves sharp edges called EC-Net \cite{yu2018ec}. 
There is another network based on \acp{GAN} \cite{goodfellow2014generative}, PU-GAN \cite{li2019pugan}. 
All these methods directly employ neural network on point cloud. Recent work PUGeo-Net \cite{qian2020pugeo} uses surface parameterization to map 3D surface to 2D parametric domain. They also estimates the normals with upsampled points. In contract to this method, our method is directly employed on point cloud.
 
Some work was also done on generating point clouds from single images such as PointSetGeneration \cite{DBLP:journals/corr/FanSG16} and DensePCR \cite{mandikal2019densepcr}.
In addition, DensePCR also upsamples the generated point cloud using multiple stages .
But all these above networks only consider point clouds as output. 
Point normals are also an important property for many applications from surface reconstruction to rendering. 
Using classical techniques, normal calculation are subjective to noise and outliers. 
In this paper, we propose to predict not only the point positions but also the corresponding normals using deep learning.

Similarly for normals prediction, as of now PCPNET \cite{GuerreroEtAl:PCPNet:EG:2018} and Nesti-Net \cite{ben_shabat2018nestinet} are existing deep learning approaches. 
PCPNET uses multi-level patches based on different scale and predicts normals. 
Similarly, Nesti-Net modifies the representation of point clouds to multi-scale point statistics (MuPS) which is then fed as input to CNN network. 
These networks only predict normals w.r.t. corresponding point clouds. 
In our work, we are upsampling the point clouds and predict normals for each point with higher accuracy.

\begin{figure*}[t]
  \center
  \begin{tikzpicture}[node distance=0, outer sep=0, inner sep=0]
      \draw (0, 0) node {\includegraphics[width=\textwidth]{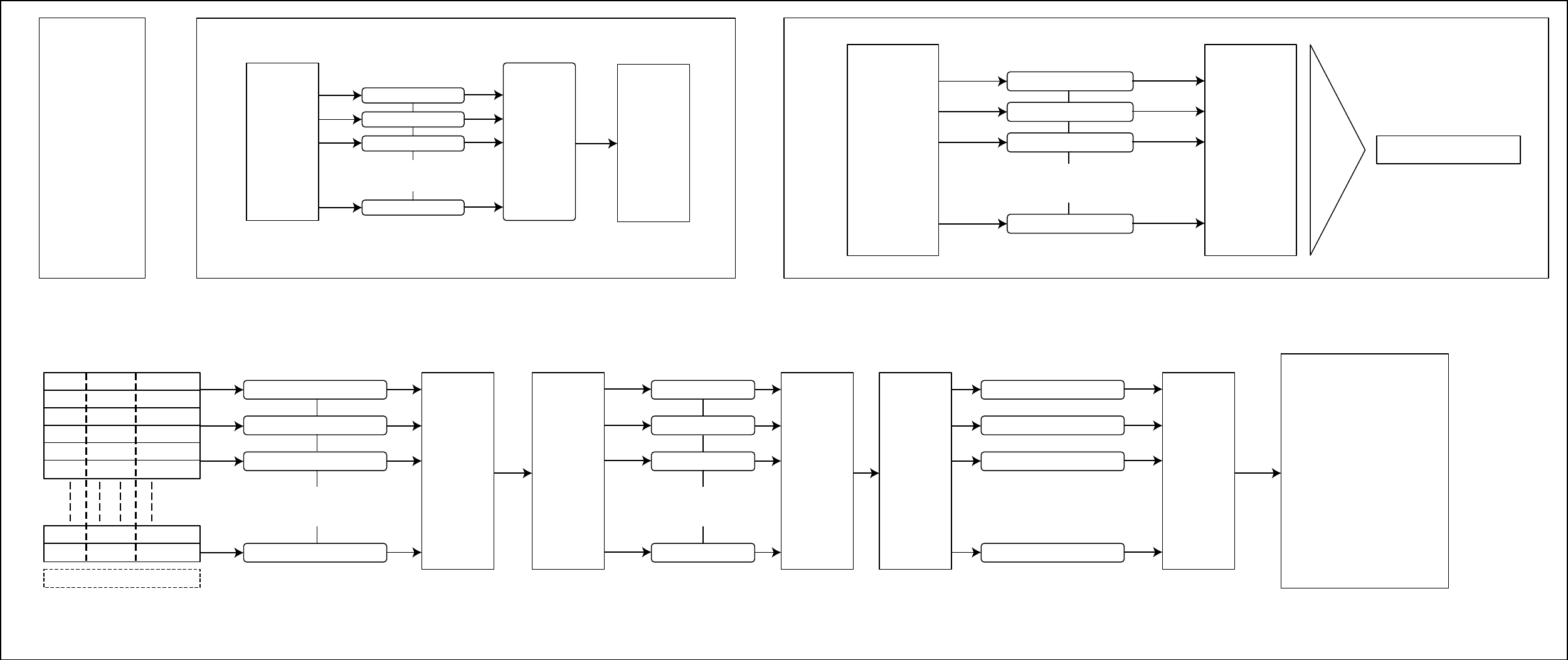}};


      \draw (-6.95, 0.25) node[scale=0.45] {i) Input points};

      \draw (-6.95, 2.6) node[scale=0.35] {$X_n$};
      \draw (-6.95, 1.9) node {\includegraphics[width=0.75cm]{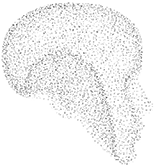}};
      \draw (-6.95, 1.1) node[scale=0.35, text width=4cm, align=center] {Input points \\ with normals \\ n x 6};


      \draw (-3.3, 0.25) node[scale=0.45] {ii) Local Feature Extraction Module};

      \draw (-5.6, 1.9) node[scale=0.35, text width=4cm, align=center, rotate=90] {Input points \\ with normals};

      \draw (-5.05, 1.9) node {\includegraphics[width=0.50cm]{assets/head_sparse.png}};
      \draw[dash pattern=on 2pt off 1pt, line width=0.08mm] (-4.97, 1.7) circle (0.08);
      \draw[dash pattern=on 2pt off 1pt, line width=0.08mm] (-4.91, 2.0) circle (0.08);
      \draw[dash pattern=on 2pt off 1pt, line width=0.08mm] (-5.2, 2.02) circle (0.08);
      \draw[dash pattern=on 2pt off 1pt, line width=0.08mm] (-5.15, 1.85) circle (0.08);
      \draw (-5.05, 1.5) node[scale=0.35] {n x 6};
      \draw (-5.05, 0.8) node[scale=0.35, text width=4cm, align=center] {Circles containing \\ neighbor points};

      \draw (-3.72, 2.6) node[scale=0.35] {mlp(32, 64, 128)};
      \draw (-3.72, 1.55) node[scale=0.35] {shared};

      \draw (-2.45, 1.9) node[scale=0.35, text width=4cm, align=center, rotate=90] {Neighborhood feature \\ max pooling};
      
      \draw (-1.3, 2.8) node[scale=0.35] {$X_l$};
      \draw (-1.3, 1.9) node[scale=0.35] {n x 512};
      \draw (-1.3, 0.8) node[scale=0.35] {local features};


      \draw (3.8, 0.25) node[scale=0.45] {iii) Global Feature Extraction Module};

      \draw (0.4, 1.8) node[scale=0.35, text width=4cm, align=center, rotate=90] {Input points \\ with normals};

      \draw (1.1, 1.8) node {\includegraphics[width=0.67cm]{assets/head_sparse.png}};

      \draw (2.88, 2.8) node[scale=0.35] {mlp(32, 64, 64, 128, 256, 512)};
      \draw (2.88, 1.5) node[scale=0.35] {shared};

      \draw (4.7, 1.8) node[scale=0.35] {n x 512};

      \draw (5.5, 1.8) node[scale=0.35, text width=1cm, align=center] {max \\ pool};

      \draw (6.7, 2.1) node[scale=0.35] {$X_g$};
      \draw (6.7, 1.8) node[scale=0.35] {512};
      \draw (6.7, 0.8) node[scale=0.35] {global features};


      \draw (-6.7, -2.9) node[scale=0.45, text width=4cm, align=center] {iv) Concatenated \\ features};

      \draw (-7.2, -0.3) node[scale=0.35] {$X_n$};
      \draw (-6.8, -0.3) node[scale=0.35] {$X_l$};
      \draw (-6.3, -0.3) node[scale=0.35] {$X_g$};

      \draw (-6.7, -2.51) node[scale=0.35] {n x (6 + 128 + 512)};


      \draw (-4.72, -0.3) node[scale=0.35] {mlp(512, 256, 128)};
      \draw (-4.72, -1.8) node[scale=0.35] {shared};


      \draw (-3.3, -1.45) node[scale=0.35] {n x 128};


      \draw (-2.1, -2.9) node[scale=0.45, text width=4cm, align=center] {v) Features \\ reshaping - I};

      \draw (-2.1, -1.45) node[scale=0.35] {2n x 64};


      \draw (-0.84, -0.3) node[scale=0.35] {mlp(64, 32)};
      \draw (-0.84, -1.8) node[scale=0.35] {shared};


      \draw (0.33, -1.45) node[scale=0.35] {2n x 32};


      \draw (1.33, -2.9) node[scale=0.45, text width=4cm, align=center] {vi) Features \\ reshaping - II};

      \draw (1.33, -1.45) node[scale=0.35] {4n x 16};


      \draw (2.7, -0.3) node[scale=0.35] {mlp(16, 6)};


      \draw (4.2, -1.45) node[scale=0.35] {4n x 6};

      
      \draw (5.9, -2.9) node[scale=0.45, text width=4cm, align=center] {vii) Dense Point Cloud \\ with normals};
      
      \draw (5.9, -1.45) node {\includegraphics[width=1.2cm]{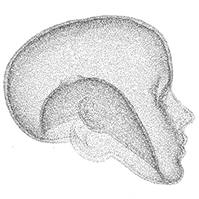}};

  \end{tikzpicture}
  \caption{Proposed network architecture for point cloud upsampling and normal estimation \textbf{i)} Input point cloud with normals \textbf{ii)} Local feature extraction \textbf{iii)} Global feature extraction \textbf{iv)} Feature concatenation \textbf{v)} Feature reshaping in points dimension by factor of 2 followed by series of shared \acsp{MLP} \textbf{vi)} Feature reshaping in points dimension by factor of 2 followed by series of shared \acsp{MLP} \textbf{vii)} Output point cloud with normals.}
  \label{fig:architecture}
 \end{figure*}

\section{\uppercase{Network Architecture}} \label{sec:network_architecture}

\noindent For a given point cloud, our deep learning architecture upsamples the applied input and predicts the point normals such that all points lie on the underlying surface with uniform distribution.
Our deep learning network includes feature embedding, feature reshaping (expansion in point dimension), and point coordinate regression. 
The transformation of the input point cloud to feature space is discussed in \ref{sec:feature_embedding}. 
Here, feature extraction and concatenation is done. 
The reshaping of features in point dimensions to upsample points is discussed in section \ref{sec:feature_reshaping}. 
After feature expansion, a series of shared \acp{MLP} with shared weights regresses the output point clouds with normals which is discussed in section \ref{sec:coordinates_regression}. 
All these components allow us to capture upsampled point clouds with point normals. 
The detailed architecture is depicted in Figure \ref{fig:architecture}.

\subsection{Feature Embedding} \label{sec:feature_embedding}

\noindent The main intent of the network is to capture the salient features of the input point cloud to learn its neighbor points. 
This is done by transforming the input points to high level vector space (called features). 
In our network, we use a combination of global and local feature extraction to better capture the properties of the point cloud.

\subsubsection{Global Features}

\noindent The knowledge of global features gives the network a better understanding of the object as a whole, e.g., features such density and spacing are model specific. 
The PointNet \cite{qi2016pointnet} architecture uses a series of shared \acp{MLP} with shared weights to extract the global features for classification and segmentation tasks. 
We also uses the same network to extract global features. 
For a input point cloud \(X_p\) with shape \(n \times 6\), the output of global features \(X_g\) is \(1 \times g\) where $g$ is the number of features extracted in the last \ac{MLP} layer.

\subsubsection{Local Features}

\noindent Global features only capture the overall property of an object but do not contain information about local properties. 
Local features are the most essential part of this network. 
The local features carry the neighborhood properties as well as surface properties such as normals. 
Local features prove to be very helpful in tasks like point cloud segmentation or surface estimation. 
We have PointNet \cite{qi2016pointnet} that extracts features per point by using a series of \acp{MLP}. 
But it does not induce local features based on distance metrics. 
An improvement to PointNet, PointNet++ \cite{qi2017pointnetplusplus} captures patches of neighborhood points for every point in the input set and performs a series of \acp{MLP} on every patch and captures local information. 
We use a series of sample and grouping modules at different scales like in PointNet++ to extract hierarchical features and combine all those information into a single feature layer. 
This is also called multi-resolution grouping. 
For an input point cloud \(X_p\) with shape \(n \times 6\), the output of local features \(X_l\) is \(n \times l\) where $l$ is the number of features extracted in the last \ac{MLP} layer of local features module.

\subsubsection{Feature Concatenation}

\noindent We concatenate all the learned features, i.e., global as well as local features with input features (i.e., input point cloud) along the feature dimension. 
The output after concatenation is \(n \times (6 + l + g)\). 
Then again we uses a series of \acp{MLP} on these concatenated features before doing feature reshaping. 
The output shape after these operation is \(n \times d\) where $d$ is multiple of the scaling factor(=\verb|up_ratio|, which is 4 in our case).

\subsection{Feature Reshaping} \label{sec:feature_reshaping}

\noindent Feature reshaping is used to expand features in point dimension space. 
Since both points and features are interchangeable this means that the learned features can also be the neighbor points.
Taking this idea, we reshape our features in point dimension to upscale the points. 
The new shape of point features after this operation is \((n\times\verb|up_ratio|) \times (d\div\verb|up_ratio|)\). 
Reshaping is always done by a factor of 2.

\subsection{Coordinates Regression} \label{sec:coordinates_regression}

\noindent After feature reshaping, we use series of \acp{MLP} layers to regress point cloud outputs with corresponding normals. 
The shape of output point cloud is \((n\times\verb|up_ratio|) \times 6\). 
The validation of the final output w.r.t. the ground truth is done by a compound loss function which is discussed in the next section.

\section{\uppercase{Loss Function}} \label{sec:loss_function}

\noindent We propose a compound loss function that consists of point and normal losses. 
The point loss is presented in section \ref{sec:point_loss} and the normal loss in section \ref{sec:normal_loss}.

\subsection{Point Loss} \label{sec:point_loss}

\noindent For point clouds, the \ac{CD} \cite{fan2017point} and \ac{EMD} \cite{rubner2000earth} are the most suitable cost functions. 
\ac{CD} tries to find the minimum distance between two sets of points, i.e., in our case the ground truth and the estimated point clouds.
In the following, let the ground truth point cloud be \( X_{p} \) and the network output be \( \widehat{X}_{p} \). 
The \ac{CD} is defined as:

\begin{equation} \label{eq:1}
\begin{split}
 d_{cd}(\widehat{X}_{p}, X_{p}) = \sum_{x \in \widehat{X}_{p}} \underset{y \in X_{p}}{min} \left \| x - y \right \|_{2}^{2} \\
 + \sum_{y \in X_{p}} \underset{x \in \widehat{X}_{p}}{min} \left \| x - y \right \|_{2}^{2} 
\end{split}
\end{equation}

\ac{EMD} tries to solve an optimization problem. 
The mathematical equation of \ac{EMD} for output \( \widehat{X}_{p} \) and ground truth \( X_{p} \)  is shown in Eq. \eqref{eq:2} where \( \phi:  \widehat{X}_{p} \rightarrow X_{p} \) is a bijection:

\begin{equation} \label{eq:2}
 d_{emd}(\widehat{X}_{p}, X_{p}) = \underset{\phi : \widehat{X}_{p} \rightarrow X_{p}}{min} \sum_{x \in \widehat{X}_{p}} \left \| x - \phi (x)) \right \|_{2}
\end{equation}

Both functions are fully differentiable.
A Comparison of both functions can be seen in \cite{fan2017point}. 
However, \ac{EMD} is more accurate than the \ac{CD}.
\ac{EMD} is more computationally intensive and requires more time and memory for high density point clouds. 
In our network, we choose \ac{CD} as a cost function for the point loss.

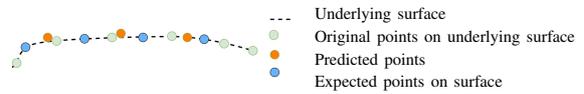
\begin{figure}[ht]
  \center
  \begin{tikzpicture}[node distance=0, outer sep=0, inner sep=0]
  \draw (0, 0) node {\includegraphics[width=0.5\textwidth, keepaspectratio]{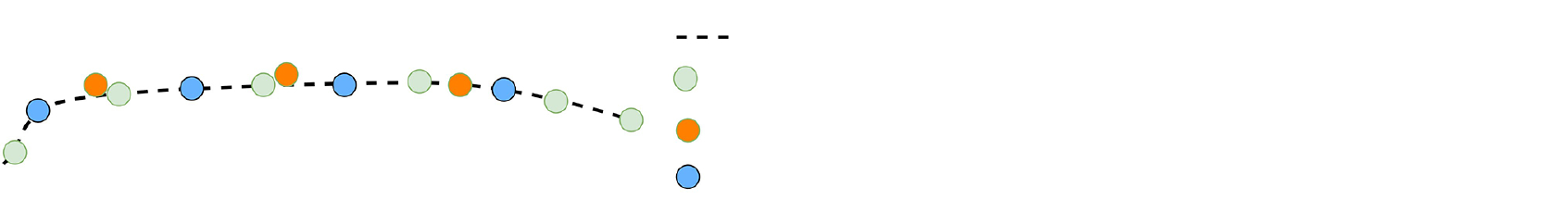}};
  \draw (2.0, 0.4) node[text width=5.6cm, scale=0.7] {\small Underlying surface};
  \draw (2.0, 0.1) node[text width=5.6cm, scale=0.7] {\small Original points on underlying surface};
  \draw (2.0, -0.2) node[text width=5.6cm, scale=0.7] {\small Predicted points};
  \draw (2.0, -0.5) node[text width=5.6cm, scale=0.7] {\small Expected points on surface};
   \end{tikzpicture}
  \caption{Non-uniform gathering of predicted points around the original points in the case of optimization using the chamfer distance \ac{CD}.}
  \label{fig:chamfer_loss}
 \end{figure}

It may happen that the predicted points lie too close to their neighbor points and this may lead to a non-uniform gathering of points in the point cloud. 
This behavior is depicted in Figure \ref{fig:chamfer_loss}

\begin{figure}[ht]
  \center
  \center
  \begin{tikzpicture}[node distance=0, outer sep=0, inner sep=0]
  \draw (0, 0) node {\includegraphics[width=0.45\textwidth, keepaspectratio ]{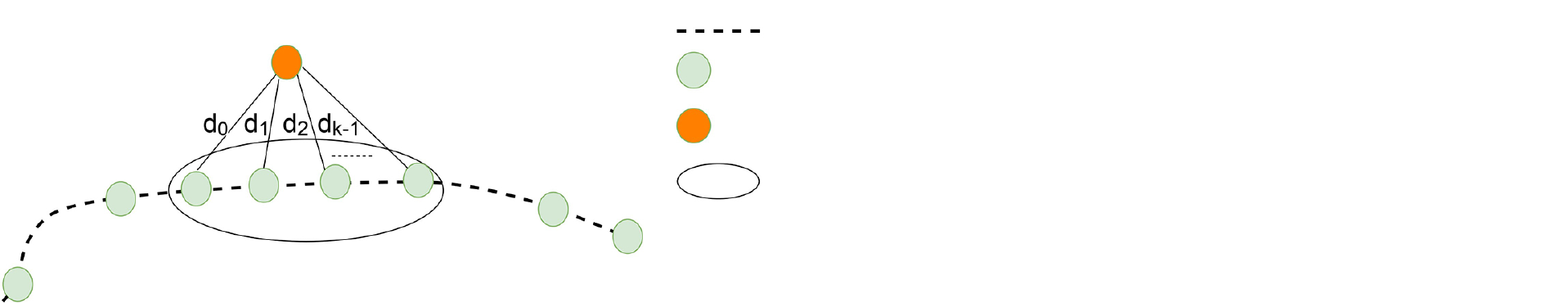}};
  \draw (2.20, 0.7) node[text width=6cm, scale=0.7] {\small Underlying surface};
  \draw (2.20, 0.4) node[text width=6cm, scale=0.7] {\small Original points on underlying  surface};
  \draw (2.20, 0.1) node[text width=6cm, scale=0.7] {\small New points};
  \draw (2.20, -0.2) node[text width=6cm, scale=0.7] {\small Containing k neighbor points $P_0$ to $P_{k-1}$};

  \draw (-1.9, -0.8) node[text width=6cm, scale=0.5] {knn points};
   \end{tikzpicture}
  \caption{Uniform distribution on underlying surface using knn point distance minimization.}
  \label{fig:unifrom_loss}
 \end{figure}
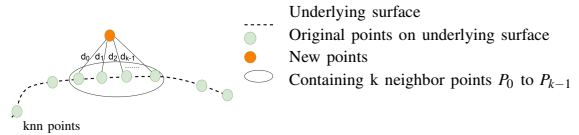

Therefore, we also include point neighbor loss as shown in Figure \ref{fig:unifrom_loss}. 
This ensures that the new points are not only closer to the ground truth but also to K nearest points in the predicted points. 
The nearest neighbor loss is shown in Eq. \eqref{eq:3}. 
For our training, we set k = 15. 

\begin{equation} \label{eq:3}
d_{point\_knn}(\widehat{X}_{p}, X_{p}) = \underset{x \in \widehat{X}_{p}}{min} \sum_{ i=0}^{k-1} \left \| x - y_{i} \right \|_{2}^{2} 
\end{equation}
\text{ \textbf{\textit{where}:} \(y_{i} \in X_{p} \) are the k nearest neighbors to x.}

Hence, the overall loss function for points is a combination of both \ac{CD} and nearest neighbor loss.


\subsection{Normal Loss} \label{sec:normal_loss}

\noindent For predicting accurate normals, we use a compound loss function too. 
For all predicted points in the upsampled point cloud, the corresponding point normals should be close to the ground truth and the adjacent points. 
This also ensures that surface reconstruction (e.g., using Poisson Reconstruction) is robust.
Hence, we use an euclidean distance metric to calculate the deviation between estimated and ground truth point normals. 
The normal loss between predicted normals \( \widehat{N}_{p} \) and ground truth normals \( N_{p} \) is shown in Eq. \eqref{eq:4}.

\begin{equation} \label{eq:4}
l_{normal}(\widehat{N}_{p}, N_{p}) = \underset{}{min} \sum_{\widehat{n} \epsilon  \widehat{N}_{p}, n \in  N_{p}} \left \| \widehat{n} - n \right \|_{2}^{2} 
\end{equation}

In addition, we need to take care that the estimated normals are orthogonal to the underlying surface. 
Since normals are perpendicular to the surface, we measure the cosine similarity which is shown in Eq. \eqref{eq:5}.

\begin{equation} \label{eq:5}
\begin{split}
l_{normal\_orth}(\widehat{X}_{p}, \widehat{N}_{p}) = min \sum_{i=0}^{k-1} \frac{\left \langle p_{l} - p_{i}, n_{l} \right \rangle}{\left \| p_{l} - p_{i} \right \|\left \| n_{l} \right \|}
\end{split}
\end{equation}
\textbf{\textit{where}:} \(p_{i}, p_{l} \in X_{p} \), \( p_{i} \) are knn to \( p_{l} \) and \( n_{l} \) normal to \( p_{l}\), $\left \langle P, N \right \rangle$ is a dot product.

Moreover, we took the assumption that for every k nearest point the normals should be the same. 
Therefore, we try to minimize this difference for each estimated normal. 
This term acts as an additional regularization on normal estimation and enforces a smooth neighborhood. 
For our training, we set k = 15. 
The loss is shown in Eq. \eqref{eq:6}.

\begin{equation} \label{eq:6}
l_{normal\_knn}(\widehat{N}_{p}) = \underset{x \in  \widehat{N}_{p}}{min} \sum_{ i=0}^{k-1} \left \| x - y_{i} \right \|_{2}^{2} 
\end{equation}
\textbf{\textit{where}:} \(y_{i} \in \widehat{N}_{p} \) are knn normals to x.



\subsubsection{Overall Loss}

\noindent The overall loss function is a combination of both point loss and normal loss. 
Since the main objective is that new points are close to the input points, more weight is given to the point \ac{CD} loss. 
The final loss function for training the network is given below with their corresponding weights. 
For our training we set $\textit{w}_1$ = 1,$\textit{w}_2$ = 0.1, $\textit{w}_3$ = 0.05, $\textit{w}_4$ = 0.0001, $\textit{w}_5$ = 0.0001. 
We set larger values for the weights of point loss as compared to normal loss. 
The weights used by the loss function are selected based on ablation study. 
We visualized the effect of different weights on the output results and thus, came up with these weights. 
The knn loss for normal shows more error in case of sharp edges thus have very small weight.

\begin{equation} \label{eq:8}
\begin{split}
loss = \textit{w}_1 d_{cd} + \textit{w}_2 d_{point\_knn} + \textit{w}_3 l_{normal} 
 \\ + \textit{w}_4 l_{normal\_orth} + \textit{w}_5 l_{normal\_knn}
\end{split}
\end{equation}

\section{\uppercase{Evaluation}} \label{sec:evaluation}

\subsection{Implementation Details and Network Training}

\noindent We implemented our network shown in Figure \ref{fig:architecture} using the PyTorch framework. 
The architecture is based on the PyTorch implementation of pointnet++\footnote{\url{https://github.com/erikwijmans/Pointnet2\_PyTorch}} for our network.
Our code is open-source and available online\footnote{\url{https://github.com/rjtshrm/point-normals-upsampling}}.

However, the input point cloud size is set to 4096 points. 
We use the Adam optimizer \cite{kingma2014adam} with a learning rate of $1e-3$ and a weight decay of $1e-5$.
The batch size is set to 20 and we train the network for 500 epochs. 
For inference, we do not feed the entire input to the network. 
We create patches of input point clouds and process each patch individually. 
Merging all upsampled patches results in very large number of points in the final result and points may be too closer in the overlapping region. 
Therefore, we use farthest point sampling to sample the merged point cloud patches to exactly four times of the size of the input point cloud.

\subsection{Datasets}

\noindent Since there are no datasets directly available for point cloud upsampling with normal inference, we create them artificially by downsampling point clouds from existing point cloud datasets.
We use the PU-Net \cite{Yu18} dataset providing point cloud patches segmented from large meshes. 
The dataset consists of 4000 patches containing 4096 points in each point cloud with corresponding normals. 
We use non-uniform downsampling to extract 1024 points from each point cloud to feed them as input to the network. 
To avoid overfitting, we apply on-the-fly data augmentation using random shifting, rotation, scaling, and adding random Gaussian noise.

\begin{figure*}[pt]
  \centering
  \begin{tikzpicture}[node distance=0, outer sep=0, inner sep=0]
  \draw (0, 0) node {\includegraphics[height=9cm, keepaspectratio]{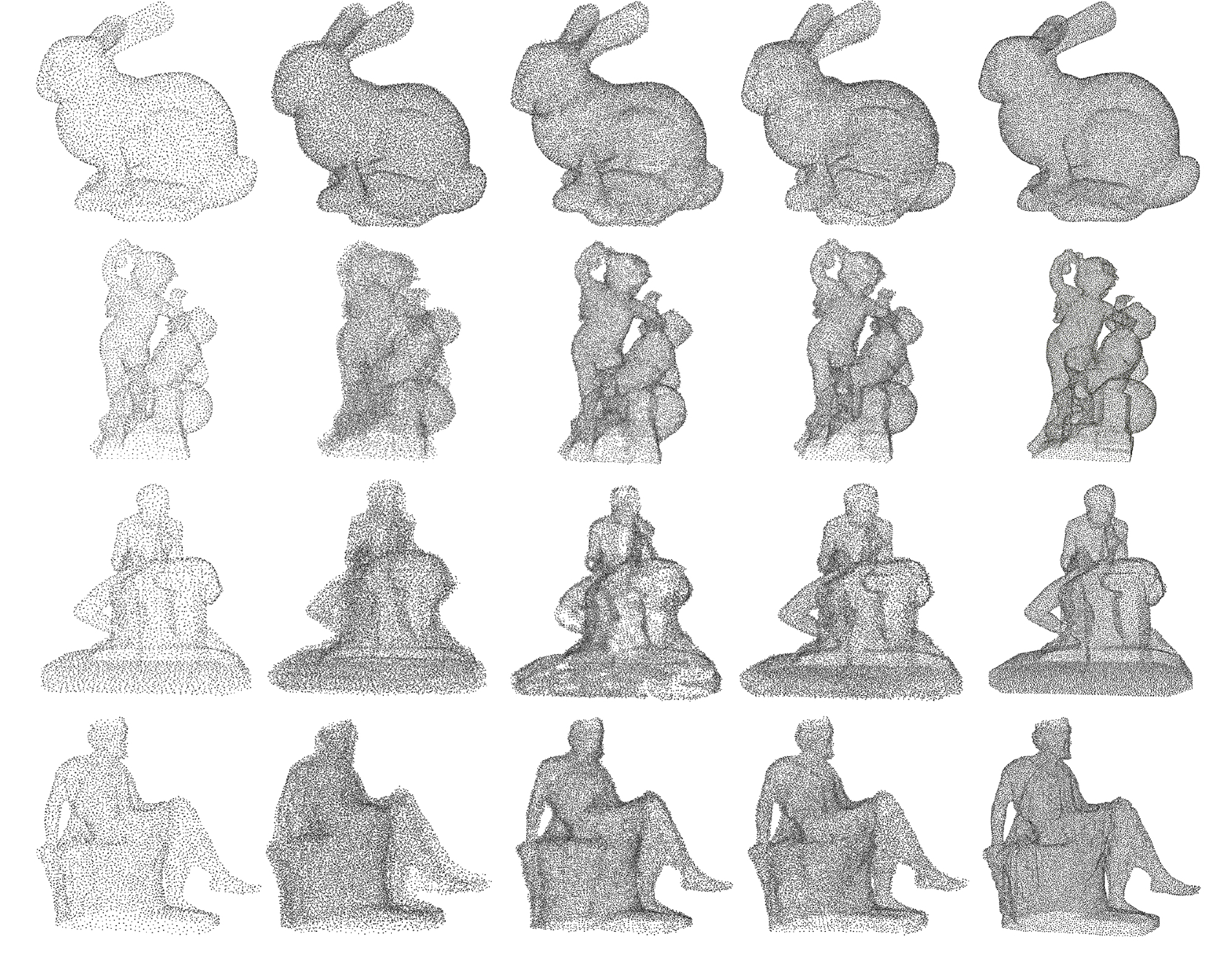}};

  \draw (-6.4, +3.4) node[scale=0.5] {pc-1};
  \draw (-6.4, +1.2) node[scale=0.5] {pc-2};
  \draw (-6.4, -1.0) node[scale=0.5] {pc-3};
  \draw (-6.4, -3.2) node[scale=0.5] {pc-4};

  \draw (-4.4, -4.5) node[scale=0.5] {i) Input};
  \draw (-2.2, -4.5) node[scale=0.5] {ii) PU-Net};
  \draw (+0.0, -4.5) node[scale=0.5] {iii) EC-Net};
  \draw (+2.2, -4.5) node[scale=0.5] {iv) Ours};
  \draw (+4.4, -4.5) node[scale=0.5] {v) Ground truth};
   \end{tikzpicture}
   
    \caption{Upsampled point clouds from different methods with input of 5000 point clouds \textbf{i)} Input \textbf{ii)} PU-Net results  \textbf{iii)} EC-Net results \textbf{iv)} Our results \textbf{v)} Ground truth }
    \label{fig:pc_comp}
    \vspace{1cm}

    \begin{tikzpicture}[node distance=0, outer sep=0, inner sep=0]
  \draw (0, 0) node {\includegraphics[height=9cm, keepaspectratio]{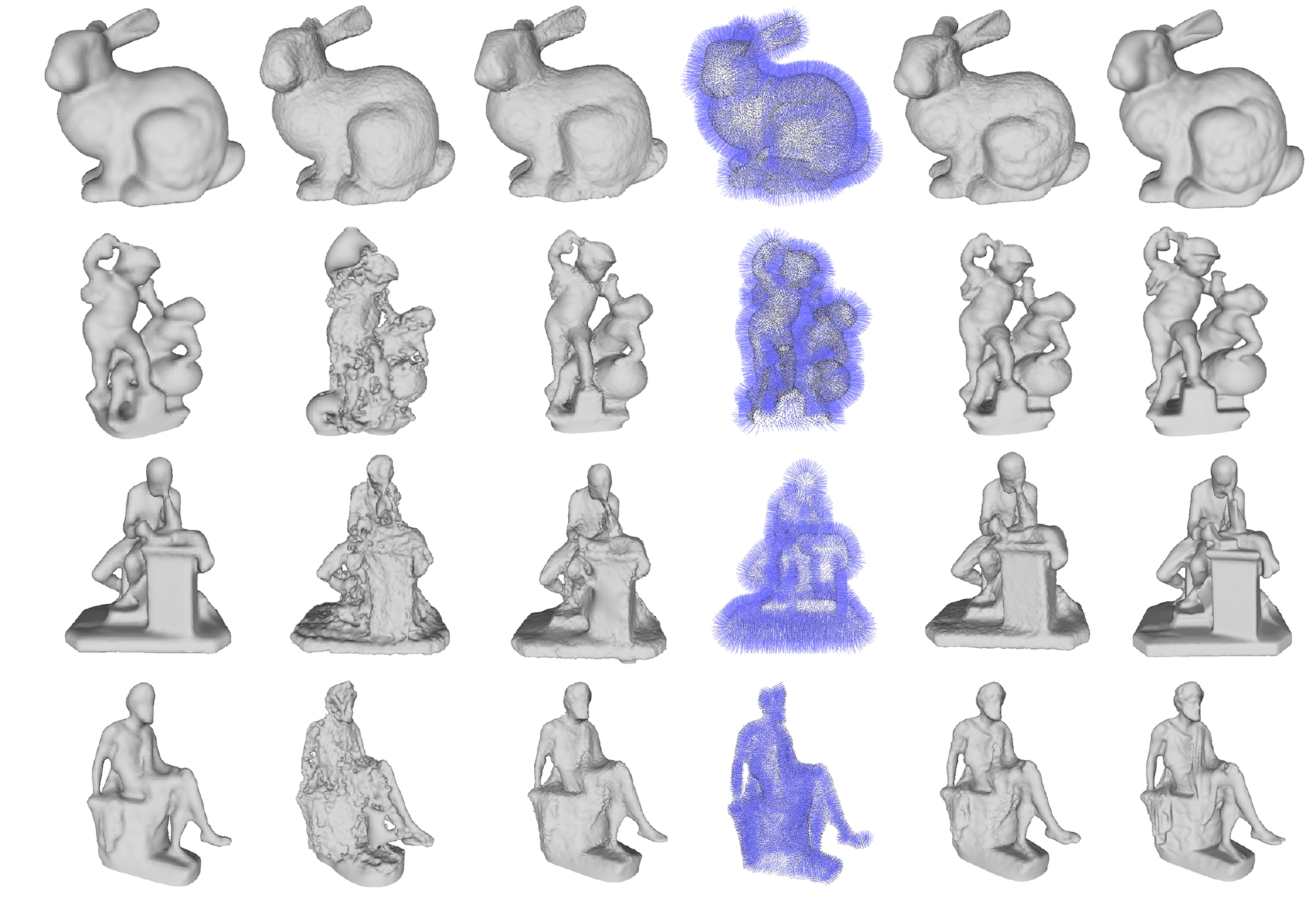}};

  \draw (-6.4, +3.4) node[scale=0.5] {pc-1};
  \draw (-6.4, +1.2) node[scale=0.5] {pc-2};
  \draw (-6.4, -1.0) node[scale=0.5] {pc-3};
  \draw (-6.4, -3.2) node[scale=0.5] {pc-4};

  \draw (-4.9, -4.5) node[scale=0.5] {i) Input};
  \draw (-2.8, -4.5) node[scale=0.5] {ii) PU-Net};
  \draw (-0.6, -4.5) node[scale=0.5] {iii) EC-Net};
  \draw (+1.4, -4.5) node[scale=0.5] {iv) Our Normals};
  \draw (+3.4, -4.5) node[scale=0.5] {v) Our reconstruction};
  \draw (+5.4, -4.5) node[scale=0.5] {vi) Ground truth};
   \end{tikzpicture}
    
    \caption{Reconstruction results from different methods with input of 5000 point clouds \textbf{i)} Input \textbf{ii)} PU-Net results  \textbf{iii)} EC-Net results \textbf{iv)} Our normals \textbf{v)} Our reconstruction \textbf{vi)} Ground truth }
    \label{fig:quality_recon}
\end{figure*}

\subsection{Results and Comparisons} \label{sec:results_and_comparison}

\noindent In this section, we will compare the results from our network w.r.t. to other upsampling networks PU-Net \cite{Yu18}, EC-Net \cite{yu2018ec}. 
We use chamfer distance \ac{CD} and \ac{HD} \cite{Berger:2013:BSR:2451236.2451246} for the evaluation of the network output. 
All methods have been tested with an upsampling ratio of 4. 
Figure \ref{fig:pc_comp} shows our results w.r.t. other approaches. 
It can be clearly seen from the figure that PU-Net results are more noisy on the surface as compared to EC-Net. 
But our results clearly outperform both PU-Net and EC-Net and are less prone to noise. 
The quantitative results for Figure \ref{fig:pc_comp} is shown in Table \ref{tab:cd_hd_table}. 
It can be infered from the table that the values of \ac{CD} and \ac{HD} for our network are small when compared to PU-Net and EC-Net. 

\begin{table}[ht]
\caption{Quantitative result with n=5000 points.}\label{tab:cd_hd_table} \centering
\begin{tabular}{cccccc}
\hline
                                              $10^{-3}$ & Method & pc-1          & pc-2           & pc-3           & pc-4           \\
\hline
\multirow{3}{*}{\rotatebox[origin=c]{90}{\acs{CD}}} & PU-Net & 0.26          & 0.22           & 0.19           & 0.15           \\
                                                    & EC-Net & 0.20          & \textbf{0.058} & 0.13           & 0.03           \\
                                                    & Ours   & \textbf{0.18} & 0.06           & \textbf{0.036} & \textbf{0.028} \\
\hline
\multirow{3}{*}{\rotatebox[origin=c]{90}{\acs{HD}}} & PU-Net & 10.63         & 10.04          & 10.14          & 8.84           \\
                                                    & EC-Net & 9.37          & 4.01           & 4.65           & 4.00           \\
                                                    & Ours   & \textbf{9.02} & \textbf{3.76}  & \textbf{4.00}  & \textbf{3.86}  \\
\hline
\end{tabular}
\end{table}

For normal evaluation, we compare the final surface reconstruction based on the predicted normals and ground truth meshes.  
We use meshlab \cite{cignoni2008meshlab} to reconstruct the point cloud to a 3D model.
The final Poisson Reconstructions for the predicted point clouds as shown in Figure \ref{fig:pc_comp} are depicted in Figure \ref{fig:quality_recon}.

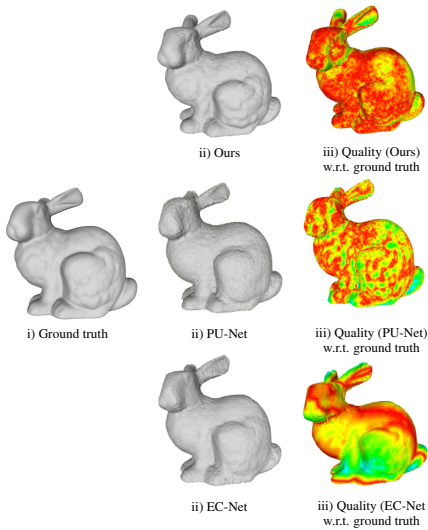
\begin{figure}[ht]
  \center
  \begin{tikzpicture}[node distance=0, outer sep=0, inner sep=0]
	  \draw (0, 0) node {\includegraphics[height=7cm]{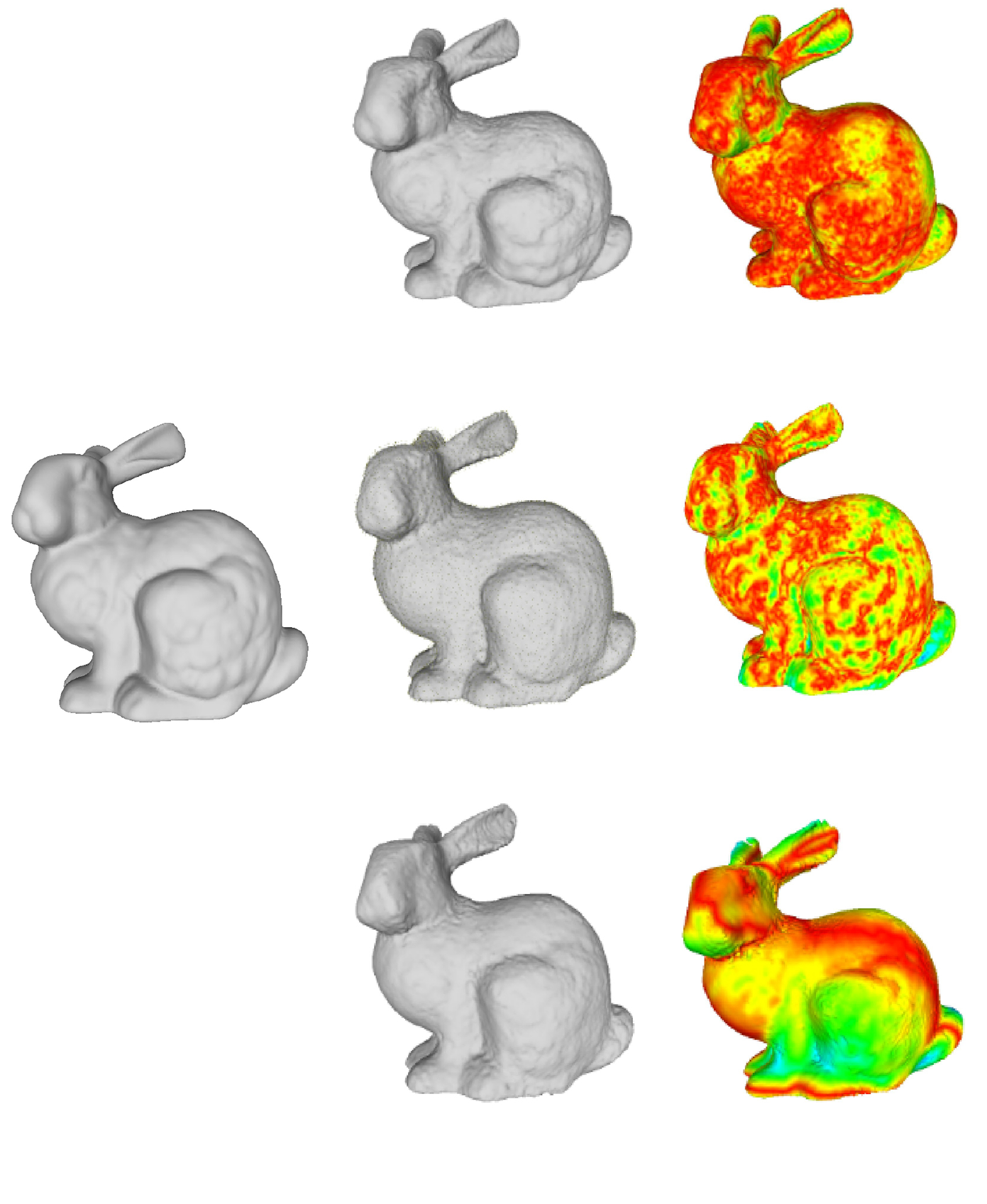}};

	  \draw (-2.0, -0.9) node[scale=0.5] {i) Ground truth};
	  \draw (0.0, -0.9) node[scale=0.5] {ii) PU-Net};
	  \draw (2.0, -0.99) node[scale=0.5, align=center] {iii) Quality (PU-Net) \\ w.r.t. ground truth};

	  \draw (0.0, 1.5) node[scale=0.5] {ii) Ours};
	  \draw (2.0, 1.41) node[scale=0.5, align=center] {iii) Quality (Ours) \\ w.r.t. ground truth};

	  \draw (0.0, -3.2) node[scale=0.5] {ii) EC-Net};
	  \draw (2.0, -3.29) node[scale=0.5, align=center] {iii) Quality (EC-Net \\ w.r.t. ground truth};
   \end{tikzpicture}
   
    \caption{Quality of our result w.r.t. ground truth for bunny dataset containing 20000 point clouds \textbf{i)} Ground truth \textbf{ii)} Our, PU-Net, EC-Net results (4x upsampling with input of 5000 point clouds) \textbf{iii)} Vertex Quality mapper (deviation from ground truth using red-green-blue color map) Red=zero error, Blue=high error.}
    \label{fig:color_mapper}
 \end{figure}
 
\begin{figure*}[pt]
  \centering
  \begin{tikzpicture}[node distance=0, outer sep=0, inner sep=0]
      \draw (0, 0) node {\includegraphics[height=9cm, keepaspectratio]{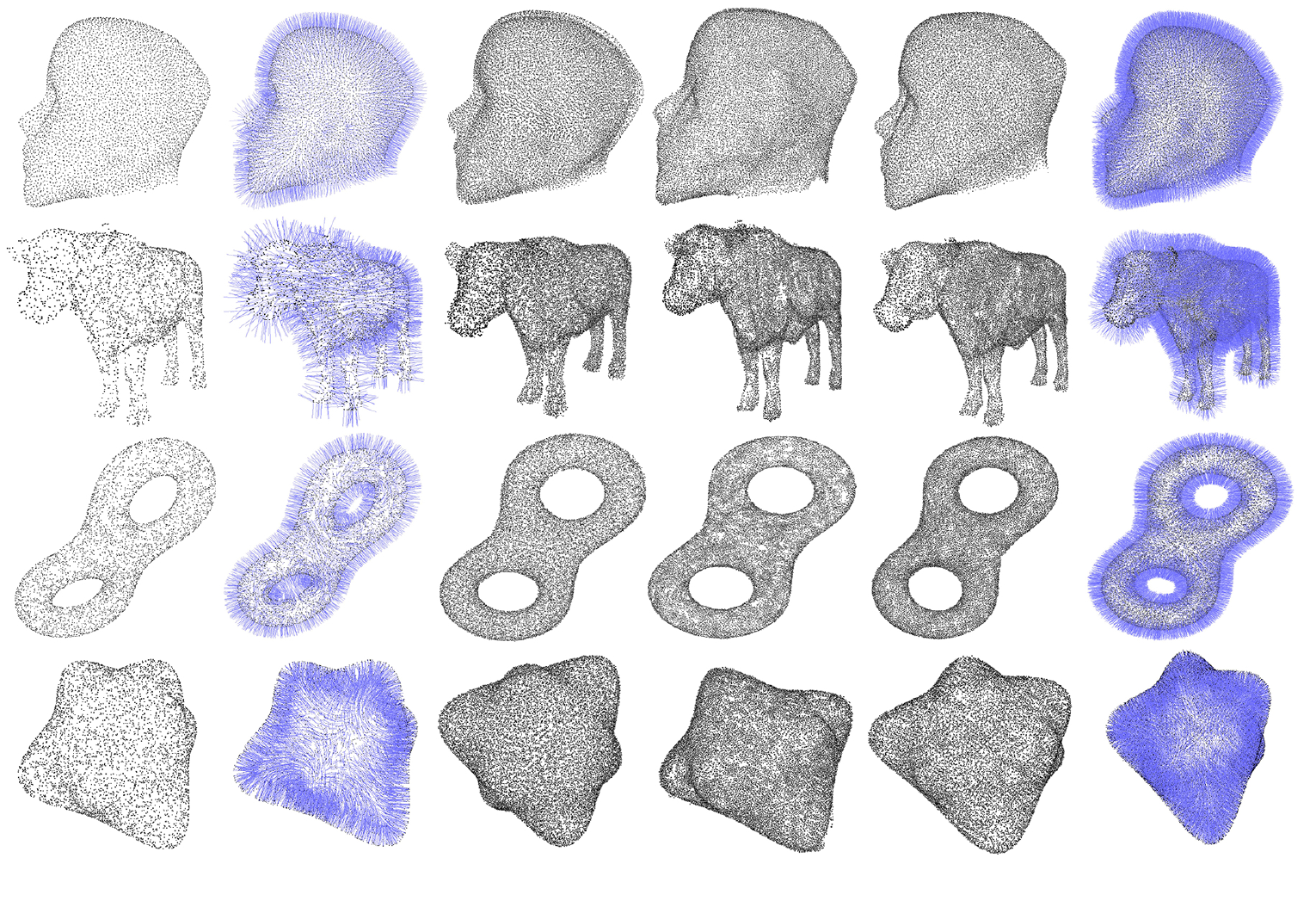}};

      \draw (-5.4, -4.5) node[scale=0.5] {i) Input};
      \draw (-3.2, -4.5) node[scale=0.5] {ii) Input normals};
      \draw (-1.2, -4.5) node[scale=0.5] {iii) PU-Net};
      \draw (+1.2, -4.5) node[scale=0.5] {iv) EC-Net};
      \draw (+3.2, -4.5) node[scale=0.5] {v) Ours};
      \draw (+5.4, -4.5) node[scale=0.5] {vi) Our with normals};
   \end{tikzpicture}
    \caption{Additional results: Upsampled point clouds from different method w.r.t. to others (Input 5000 points and 4x upsampled output) \textbf{i)} Input point cloud \textbf{ii)} Input normals \textbf{iii)} PU-Net results \textbf{iv)} EC-Net results \textbf{v)} Our results (point cloud) \textbf{vi)} Our results (point cloud normals)}
    \label{fig:comparison}

    \vspace{1cm}

    \begin{tikzpicture}[node distance=0, outer sep=0, inner sep=0]
      \draw (0, 0) node {\includegraphics[height=9cm, keepaspectratio]{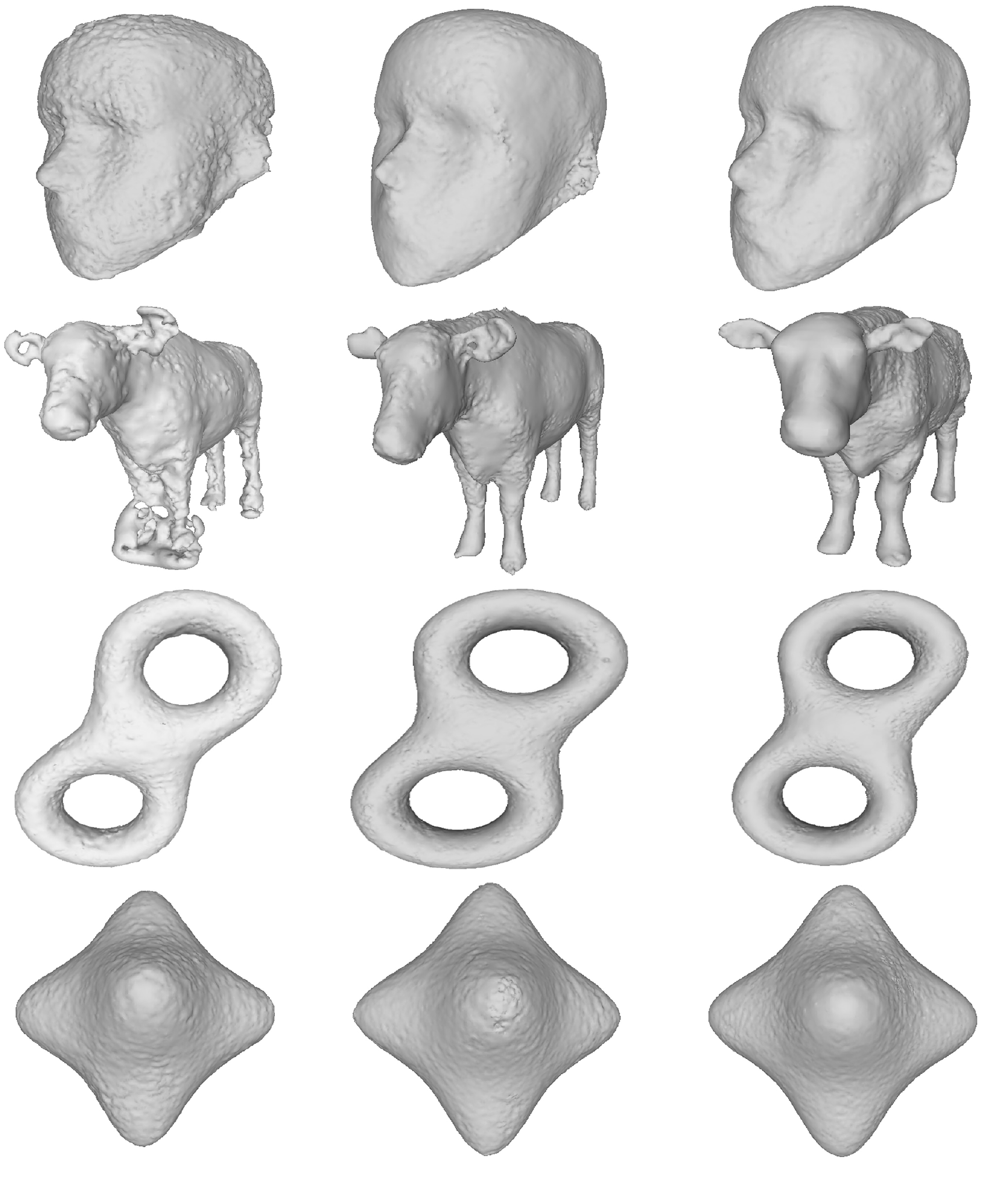}};

      \draw (-2.5, -4.5) node[scale=0.5] {i) PU-Net};
      \draw (0.0, -4.5) node[scale=0.5] {ii) EC-Net};
      \draw (2.5, -4.5) node[scale=0.5] {iii) Ours};
   \end{tikzpicture}
    \caption{Additional results: Surface reconstruction result comparisons of different architecture w.r.t. to others (Input 5000 points and 4x upsampled output) \textbf{i)} PU-Net \textbf{ii)} EC-Net \textbf{iii)} Ours }
     \label{fig:recon}
 \end{figure*}

Figure \ref{fig:color_mapper} shows the deviation of the final mesh compared to the ground truth using RGB color coding. 
Red color represents the zero error, blue represents maximum error and the transition from red to blue has increasing error values. 
It can be seen that our results for the bunny dataset are better when compared to PU-Net and EC-Net. 
Figure \ref{fig:comparison} depicts some additional results of our method compared to state-of-the-art approaches. 
In the first row (head point cloud), PU-NET creates multiple layers on the surface which leads to wrong surface reconstructions. 
Similarly EC-NET estimates a point cloud that is more noisy than ours.
Unlike these networks, our network also predicts the normals which can be seen in last column Figure \ref{fig:comparison}. 
The reconstruction based on these normals is shown in Figure \ref{fig:recon}.

\section{\uppercase{Discussion and Limitations}} \label{sec:discussion}
 
\noindent Point clouds are an important representation for 3D objects. 
They have a wide range of applications from visualization and rendering to surface reconstruction.
These applications also consider surface normals in addition to the points. 
Therefore, our approach additionally estimates surface normals along with the upsampled point cloud.
We showed that both global features as well local features are required to learn the geometry of an input point cloud.
The output point cloud based on our feature extraction and concatenation is more accurate and estimated surface normals are mostly oriented in the correct direction. 
It can be seen from the results that our method predicts more consistent and less noisy point clouds compared to the state of the art (PU-Net and EC-Net).
The final surface reconstruction based on the output point normals is also better.

\begin{figure}[h]
  \center
  \begin{tikzpicture}[node distance=0, outer sep=0, inner sep=0]
  \draw (0, 0) node {\includegraphics[width=\linewidth, keepaspectratio]{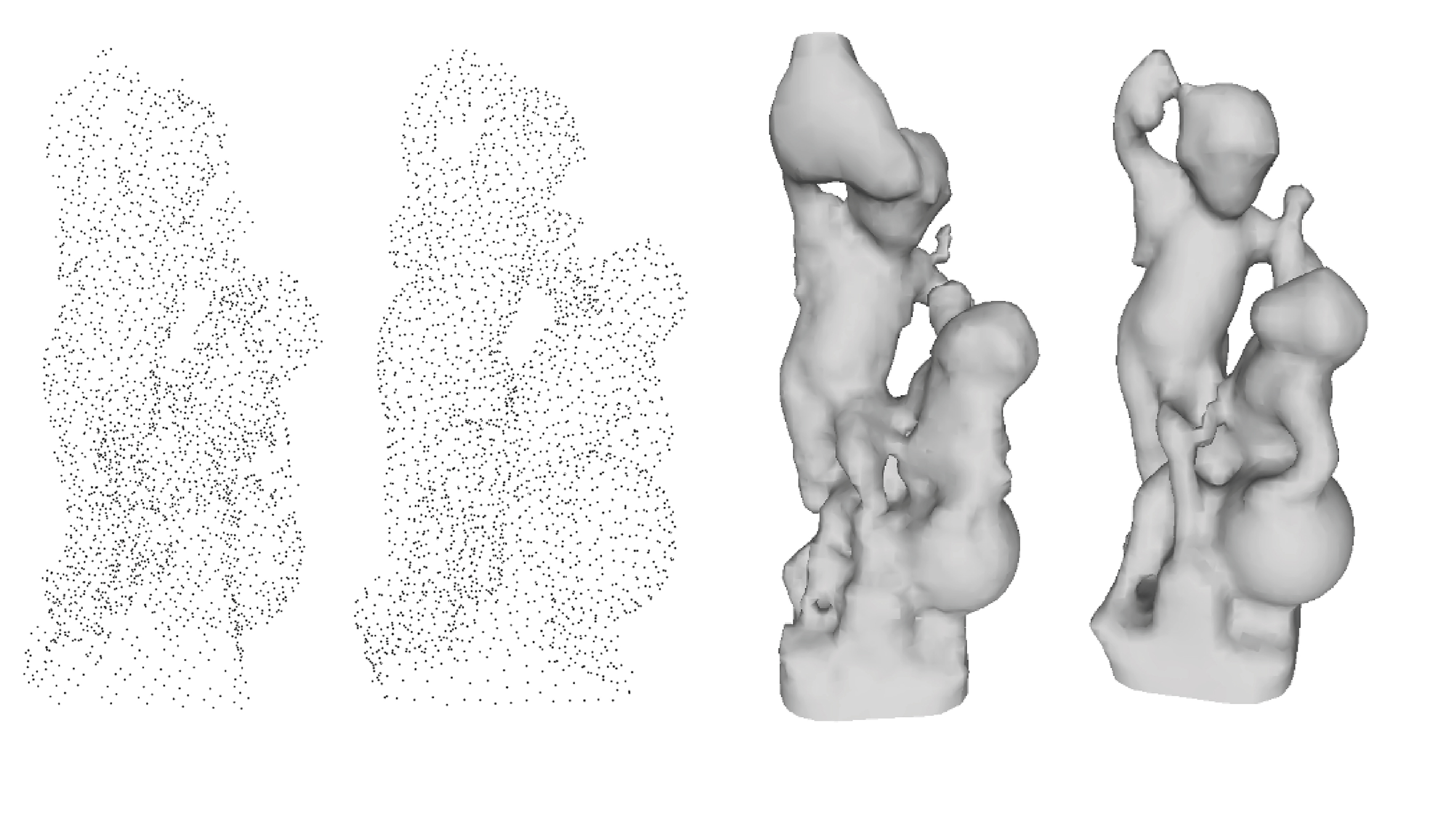}};

  \draw (-3.0, -2) node[scale=0.5] {i) Ours};
  \draw (-1.0, -2) node[scale=0.5] {ii) GT};
  \draw (0.8, -2) node[scale=0.5] {iii) Our reconstruction};
  \draw (2.6, -2) node[scale=0.5] {iv) GT reconstruction};
   \end{tikzpicture}
   
    \caption{Upsampling results of sparse point cloud (input = 625 points and output = 25000 points).}
    \label{fig:sparse_result}
 \end{figure}
 
But our network also has some limitations.
One of the limitations is that our network does not work well on very sparse inputs.
Our results can be seen in Figure \ref{fig:sparse_result} w.r.t. to the ground truth. 
The normals predicted for sparse inputs are also not oriented in the right direction which is why the final reconstruction is also distorted.
We also apply a patch based approach for the inference and this may result in misalignments at the patch borders.
Hence, the Poisson Reconstruction is not perfectly smooth in areas where patches overlap.
Also, our network is not capable of shape completion, i.e., it does not synthesize missing parts in shapes.
Our main focus in this paper was to upsample point clouds. 
Shape completion is a separate problem, requiring further research.

\section{\uppercase{Conclusion}} \label{sec:conclusion}

\noindent In this paper, we presented a deep learning network to not only upsample the point cloud but also to estimate corresponding surface normals. 
Our network uses input features, local, and global features to learn the final estimation of point cloud data.
We introduced feature reshaping in point dimension to upsample the point clouds.
We also introduced a compound loss function combining point and normal loss. 
It ensures that new points are on underlying surfaces and the normals estimated are orthogonal to surfaces with correct orientations. 
Moreover, we showed that predicting normals results in smoother and more consistent upsampled point clouds. 
This is especially beneficial for tasks like surface reconstruction.
In this work, we focused on raw point clouds with normals. 
In future work we will also extend the network to include colored point clouds and estimate the color for each upsampled point. 
Another direction of future work is looking into shape completion to fill in holes in the input point cloud, which often result from real world scans.



\section*{\uppercase{Acknowledgements}}
The underlying research of these results has been partially funded by the Free State of Thuringia with the number \textbf{2015 FE 9108} and co-financed by the European Union as part of the European Regional Development Fund (ERDF).


%
%
\bibliographystyle{apalike}
{\small
\bibliography{bib}}

\begin{thebibliography}{}

\bibitem[Ben-Shabat et~al., 2018]{ben_shabat2018nestinet}
Ben-Shabat, Y., Lindenbaum, M., and Fischer, A. (2018).
\newblock Nesti-net: Normal estimation for unstructured 3d point clouds using
  convolutional neural networks.
\newblock {\em arXiv preprint arXiv:1812.00709}.

\bibitem[Berger et~al., 2013]{Berger:2013:BSR:2451236.2451246}
Berger, M., Levine, J.~A., Nonato, L.~G., Taubin, G., and Silva, C.~T. (2013).
\newblock A benchmark for surface reconstruction.
\newblock {\em ACM Trans. Graph.}, 32(2):20:1--20:17.

\bibitem[Cignoni et~al., 2008]{cignoni2008meshlab}
Cignoni, P., Callieri, M., Corsini, M., Dellepiane, M., Ganovelli, F., and
  Ranzuglia, G. (2008).
\newblock {MeshLab: an Open-Source Mesh Processing Tool}.
\newblock In Scarano, V., Chiara, R.~D., and Erra, U., editors, {\em
  Eurographics Italian Chapter Conference}. The Eurographics Association.

\bibitem[Fan et~al., 2016]{DBLP:journals/corr/FanSG16}
Fan, H., Su, H., and Guibas, L.~J. (2016).
\newblock A point set generation network for 3d object reconstruction from a
  single image.
\newblock {\em CoRR}, abs/1612.00603.

\bibitem[Fan et~al., 2017]{fan2017point}
Fan, H., Su, H., and Guibas, L.~J. (2017).
\newblock A point set generation network for 3d object reconstruction from a
  single image.
\newblock In {\em Proceedings of the IEEE conference on computer vision and
  pattern recognition}, pages 605--613.

\bibitem[Goodfellow et~al., 2014]{goodfellow2014generative}
Goodfellow, I., Pouget-Abadie, J., Mirza, M., Xu, B., Warde-Farley, D., Ozair,
  S., Courville, A., and Bengio, Y. (2014).
\newblock Generative adversarial nets.
\newblock In {\em Advances in neural information processing systems}, pages
  2672--2680.

\bibitem[Guerrero et~al., 2018]{GuerreroEtAl:PCPNet:EG:2018}
Guerrero, P., Kleiman, Y., Ovsjanikov, M., and Mitra, N.~J. (2018).
\newblock {PCPNet}: Learning local shape properties from raw point clouds.
\newblock {\em Computer Graphics Forum}, 37(2):75--85.

\bibitem[H{\"a}ne et~al., 2019]{hspHane19}
H{\"a}ne, C., Tulsiani, S., and Malik, J. (2019).
\newblock Hierarchical surface prediction.
\newblock {\em TPAMI}.

\bibitem[Huang et~al., 2009]{Huang2009ConsolidationReconstruction}
Huang, H., Li, D., Zhang, H., Ascher, U., and Cohen-Or, D. (2009).
\newblock {Consolidation of unorganized point clouds for surface
  reconstruction}.
\newblock {\em ACM SIGGRAPH Asia 2009 papers on - SIGGRAPH Asia '09}, page~1.

\bibitem[Huang et~al., 2013]{Huang2013Edge-awareResampling}
Huang, H., Wu, S., Gong, M., Cohen-Or, D., Ascher, U., and Zhang, H.~R. (2013).
\newblock {Edge-aware point set resampling}.
\newblock {\em ACM Transactions on Graphics}, 32(1):1--12.

\bibitem[Kingma and Ba, 2014]{kingma2014adam}
Kingma, D.~P. and Ba, J. (2014).
\newblock Adam: A method for stochastic optimization.
\newblock {\em arXiv preprint arXiv:1412.6980}.

\bibitem[Li et~al., 2019]{li2019pugan}
Li, R., Li, X., Fu, C.-W., Cohen-Or, D., and Heng, P.-A. (2019).
\newblock Pu-gan: a point cloud upsampling adversarial network.
\newblock In {\em {IEEE} International Conference on Computer Vision ({ICCV})}.

\bibitem[Lipman et~al., 2007]{Lipman2007Parameterization-freeReconstruction}
Lipman, Y., Cohen-Or, D., Levin, D., and Tal-Ezer, H. (2007).
\newblock {Parameterization-free projection for geometry reconstruction}.
\newblock {\em ACM Transactions on Graphics}, 26(3):22.

\bibitem[Mandikal and Babu, 2019]{mandikal2019densepcr}
Mandikal, P. and Babu, R.~V. (2019).
\newblock Dense 3d point cloud reconstruction using a deep pyramid network.
\newblock In {\em Winter Conference on Applications of Computer Vision
  ({WACV})}.

\bibitem[Maturana and Scherer, 2015]{maturana2015voxnet}
Maturana, D. and Scherer, S. (2015).
\newblock Voxnet: A 3d convolutional neural network for real-time object
  recognition.
\newblock In {\em 2015 IEEE/RSJ International Conference on Intelligent Robots
  and Systems (IROS)}, pages 922--928. IEEE.

\bibitem[Qi et~al., 2016]{qi2016pointnet}
Qi, C.~R., Su, H., Mo, K., and Guibas, L.~J. (2016).
\newblock Pointnet: Deep learning on point sets for 3d classification and
  segmentation.
\newblock {\em arXiv preprint arXiv:1612.00593}.

\bibitem[Qi et~al., 2017]{qi2017pointnetplusplus}
Qi, C.~R., Yi, L., Su, H., and Guibas, L.~J. (2017).
\newblock Pointnet++: Deep hierarchical feature learning on point sets in a
  metric space.
\newblock In Guyon, I., Luxburg, U.~V., Bengio, S., Wallach, H., Fergus, R.,
  Vishwanathan, S., and Garnett, R., editors, {\em Advances in Neural
  Information Processing Systems 30}, pages 5099--5108. Curran Associates, Inc.

\bibitem[Qian et~al., 2020]{qian2020pugeo}
Qian, Y., Hou, J., Kwong, S., and He, Y. (2020).
\newblock Pugeo-net: A geometry-centric network for 3d point cloud upsampling.
\newblock {\em arXiv}, pages arXiv--2002.

\bibitem[Riegler et~al., 2017a]{Riegler2017OctNetFusion}
Riegler, G., Ulusoy, A.~O., Bischof, H., and Geiger, A. (2017a).
\newblock Octnetfusion: Learning depth fusion from data.
\newblock In {\em Proceedings of the International Conference on 3D Vision}.

\bibitem[Riegler et~al., 2017b]{Riegler2017OctNet}
Riegler, G., Ulusoy, A.~O., and Geiger, A. (2017b).
\newblock Octnet: Learning deep 3d representations at high resolutions.
\newblock In {\em Proceedings of the IEEE Conference on Computer Vision and
  Pattern Recognition}.

\bibitem[Rubner et~al., 2000]{rubner2000earth}
Rubner, Y., Tomasi, C., and Guibas, L.~J. (2000).
\newblock The earth mover's distance as a metric for image retrieval.
\newblock {\em International journal of computer vision}, 40(2):99--121.

\bibitem[Sun et~al., 2015]{Sun2015DenoisingMinimization}
Sun, Y., Schaefer, S., and Wang, W. (2015).
\newblock {Denoising point sets via L0 minimization}.
\newblock {\em Computer Aided Geometric Design}, 35-36:2--15.

\bibitem[Wang et~al., 2019]{dgcnn}
Wang, Y., Sun, Y., Liu, Z., Sarma, S.~E., Bronstein, M.~M., and Solomon, J.~M.
  (2019).
\newblock Dynamic graph cnn for learning on point clouds.
\newblock {\em ACM Transactions on Graphics (TOG)}.

\bibitem[Weinmann, 2016]{weinmann2016reconstruction}
Weinmann, M. (2016).
\newblock {\em Reconstruction and Analysis of 3D Scenes: From Irregularly
  Distributed 3D Points to Object Classes}.
\newblock Springer Publishing Company, Incorporated, 1st edition.

\bibitem[Yu et~al., 2018a]{yu2018ec}
Yu, L., Li, X., Fu, C.-W., Cohen-Or, D., and Heng, P.-A. (2018a).
\newblock Ec-net: an edge-aware point set consolidation network.
\newblock In {\em ECCV}.

\bibitem[Yu et~al., 2018b]{Yu18}
Yu, L., Li, X., Fu, C.-W., Cohen-Or, D., and Heng, P.-A. (2018b).
\newblock Pu-net: Point cloud upsampling network.
\newblock In {\em Proceedings of IEEE Conference on Computer Vision and Pattern
  Recognition (CVPR)}.

\bibitem[Zhang and Rabbat, 2018]{ZhangR_18_gcnn_point_cloud}
Zhang, Y. and Rabbat, M. (2018).
\newblock A graph-cnn for 3d point cloud classification.
\newblock In {\em International Conference on Acoustics, Speech and Signal
  Processing (ICASSP)}, Calgary, Canada.

\end{thebibliography}

\end{document}